\documentclass[journal,twoside,web]{ieeecolor}
\usepackage{jsen}
\usepackage{cite}
\usepackage{amsmath,amssymb,amsfonts}

\usepackage{graphicx}
\usepackage{array}
\usepackage{textcomp}
\usepackage{wrapfig}
\usepackage{multirow}  
\usepackage{tabularx}
\usepackage{subcaption}
\usepackage{colortbl}
\definecolor{LG}{rgb}{0.5,1,0.5}
\definecolor{Yel}{rgb}{1,1,0}
\usepackage{hyperref}
\usepackage{algorithm}
\usepackage{algpseudocode}
\makeatletter
\renewcommand{\ALG@name}{Procedure}
\makeatother

\usepackage{xcolor}
\usepackage{optidef}

\usepackage{pifont}% http://ctan.org/pkg/pifont

\def\BibTeX{{\rm B\kern-.05em{\sc i\kern-.025em b}\kern-.08em
    T\kern-.1667em\lower.7ex\hbox{E}\kern-.125emX}}
\definecolor{abstractbg}{rgb}{0.89804,0.94510,0.83137}
\begin{document}
\title{Filling the Pareto optimal front for affordance segmentation on embedded devices using RGB-D cameras\\ \thanks{Project funded under the National Recovery and Resilience Plan (NRRP), Mission 4 Component 2 Investment 1.1 - Call for tender No. 1049 published on Sept 14, 2022 by the Italian Ministry of University and Research (MUR) funded by the European Union – NextGenerationEU. Project Title "LEARN - muLtimodal Edge computing-bAsed weaRable exoskeletoNs for assistance in daily life" – CUP D53D23016190001 - Grant Assignment Decree No. 1181 adopted on July 27, 2023 by MUR. \includegraphics[width=0.48\textwidth]{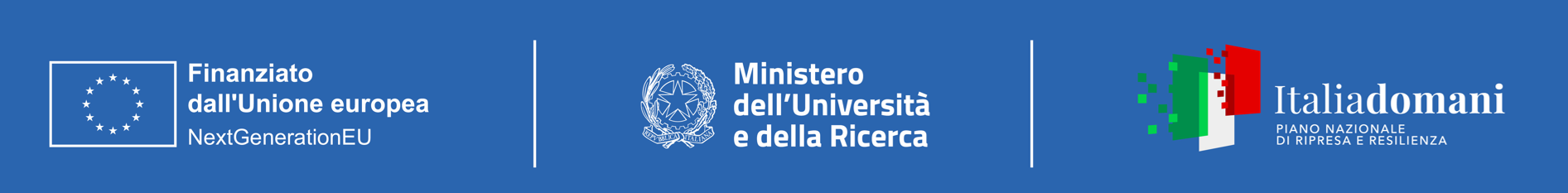}}}
\author{Edoardo Ragusa~\IEEEmembership{(Member,~IEEE),} Giovanni Paolo Canuti, Simone Lugani, Rodolfo Zunino, and Paolo Gastaldo  
\thanks{E. Ragusa, G.P. Canuti, S. Lugani, R. Zunino, and P. Gastaldo are with DITEN, University of Genoa, Genova, Italy e-mail: edoardo.ragusa@unige.it}}

\IEEEtitleabstractindextext{%
\fcolorbox{abstractbg}{abstractbg}{%
\begin{minipage}{\textwidth}%
\begin{wrapfigure}[22]{r}{3in}%
\includegraphics[width=3in]{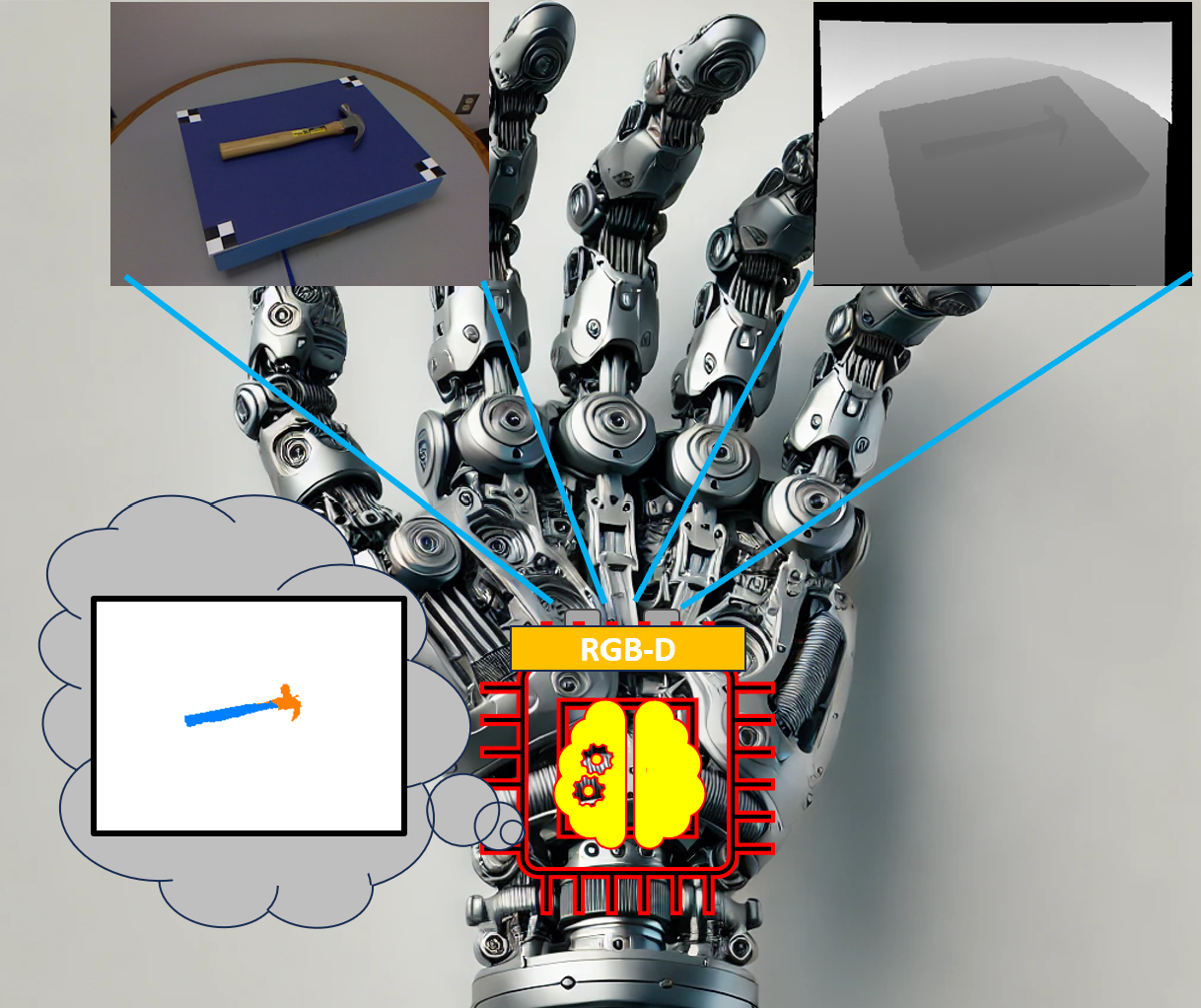}%
\end{wrapfigure}%
\begin{abstract}

While depth sensors have the potential to complement RGB data for affordance segmentation in wearable robots, their usage seems to remain underexplored.
The paper proposes two approaches: a reformulated version of hardware-aware neural architecture search,  endowed with a newly designed search space to integrate depth (D) information into small-sized deep networks, and a dedicated fine-tuning approach, including a preprocessing layer to merge depth information with RGB data and make it compatible with conventional architectures. 
In both cases, those methods aim to generate solutions that benefit from modern (portable) hardware accelerators and overcome existing tiny-like approaches, which often fail to tackle critical scenarios due to the severe constraints set by the supporting hardware. 

Extensive experiments on a pair of real-world datasets demonstrate the effectiveness of the proposed method as compared with existing solutions. The approach presented in the paper generates, in most cases, solutions that identify the Pareto optimal front to balance generalization performance and hardware requirements. The paper also describes the supporting prototype, including a Jetson Nano board and a RealSense RGB-D camera. When considering the energy profile of the device, the overall system can attain real-time performances within an energy budget that is compatible with standard batteries, such as those used in smartphones. 
\end{abstract}

\end{minipage}}}

\maketitle
%%%%%%%%%%%%%%%%%%%%%%%%%%%%%%%%%%%%%%%%%%%%%%%%%%%%%%%%%%%%%%%%%%%%%%%%%%%%%%%%%%%%%%%%%%%%%%%%%%%%%%%%%%%%%%%%%%%%%%%%%%%%%%%%%%%
\section{Introduction}

Wearable robots are essential in a variety of applications, ranging from prosthetics to those industrial settings in which robotic arms can take over strenuous or hazardous tasks for humans \cite{tang2022wearable}. Semi-autonomous control is a critical feature for these robots, as it enables the automation of certain aspects of the control process, and relieves the user of those chores when facing complex tasks \cite{salminger2022current}. For instance, a user can aim a robotic hand at an object and activate the device; the system will autonomously perform the grasping action and reduce user effort, thus enhancing the overall user's experience \cite{tang2022wearable, sun2020real}.

At the same time, wearable robots need advanced sensing capabilities to implement semi-autonomous features \cite{chiariotti2024future}: for instance, non-contact sensing is essential to the semi-autonomous control of grasping actions  \cite{krausz2019survey}; cameras and depth sensors typically serve as primary input devices, though radars have also been successfully adopted \cite{mastinu2024explorations}. Processing the captured signals on-site allows to extract the high-level information that is crucial in control pipelines. 

Affordance Segmentation (AS) is a well-known input method for such control pipelines. It splits an object into its functional parts, and enables to apply fine-grained control strategies even when handling objects with complex geometries \cite{castro2022continuous,ragusa2021hardware}. Deep Neural Networks (DNNs) seem to be the only viable approach to support the complexity of AS \cite{chu2019learning, jiang2021synergies, khalifa2022towards}; the deployment of DNNs on portable devices, however, poses a major challenge, as weight, size, and energy consumption of the embedded electronic system must be minimized \cite{ragusa2023affordance}. 

This paper focuses on a design strategy for efficient near-sensor inference when using state-of-the-art sensor technology; in particular, the paper proposes a design strategy yielding DNNs for AS using RGB-D data. Depth sensors provide a bulky, yet essential tool for an effective control pipeline \cite{castro2022continuous}, since the geometry information provided by depth data can provide insights about the shape of an observed object \cite{lopes2022survey}. Preliminary studies about the use of depth information for AS \cite{ragusa2023affordanceAP} showed that the contribution of raw depth sensing could result marginal, if one mixed depth and  RGB data in a straightforward manner. That research, at the same time, hinted at the fact that depth information could indeed improve AS in the presence of specific treatment. 

Depth information processing calls for a dedicated processing chain thus increasing computational complexity. To cope with that issue, hardware accelerators for DNNs are continuously improving at a fast rate. DNNs can now run in the presence of strict constraints on a system's size and overall power budget \cite{akkad2023embedded, saha2022machine}. Current solutions mainly rely on either the tinyML paradigm or high-performance accelerators. The design of mid-range, application-specific DNN architectures seems to be left rather unexplored, in spite of the fact that those architectures can indeed address real-life scenarios effectively. Basic advantages consist in resilience to variations in framing, background, and illumination, while maintaining acceptable hardware requirements for portable devices \cite{ragusa2023affordance}.

The approach proposed in this paper aims to benefit from depth sensors while balancing hardware requirements and robustness. The main contribution is a hardware-aware neural architecture search (HW-NAS) strategy, featuring a new search space optimized for RGB-D inputs. The paper also presents an additional  solution, based on a fine-tuning process to extend pre-trained networks on RGB data to handle RGB-D inputs.  

Experimental results validated the approach on well-known benchmarks, namely, UMD \cite{myers2015affordance}  and IIT \cite{nguyen2016detecting}. The resulting architecture lay consistently on the Pareto optimal front, when considering its generalization performances versus both model size and FLOPs. The method always outperformed existing hardware-efficient approaches, and the resulting models were successfully deployed on a Jetson Nano board, interfaced with a RealSense RGB-D camera, to verify real-time inference.

The main contributions of the presented research consist in:
\begin{itemize}
    \item A smart device for extracting high-level semantic information directly from RGB-D cameras. Tuning the DNN architecture to the available sensing and computing resources supported AS capabilities on the test prototype, featuring a limited power consumption and real-time performances.
    \item A novel HW-NAS framework for the design of lightweight DNNs with RGB-D input. The involved  search space covers RGB-D data and scales better than existing HW-NAS solutions when hardware constraints are relaxed. The approach enables multi-resolution feature extraction and supports multi-branch neural network architectures; as a result, it allows to merge different input sources while learning optimal structures directly from training data.

    \item A fast, effective solution for reusing networks (pre-trained on RGB data) to cover depth information. The integrated approach enhances generalization ability with a negligible impact on FLOPs and network parameters.
    
%    \item The proposed search space is also tested on standard RGB data, demonstrating improved performance in balancing generalization and hardware requirements, thanks to the extensive use of skip connections, which facilitate the preservation of geometric information.
    
    \item A set of DNNs that are Pareto-optimal in the generalization performance versus hardware requirements, as confirmed by results on established real-world benchmarks.

\end{itemize}

\section{Related Works}\label{sec:related}

\subsection{Affordance Segmentation for Wearable Robots}
The literature includes various prototypes and approaches using teleceptive sensing devices \cite{hundhausen2019resource, starke2022semi}, such as RGB cameras \cite{weiner2022designing}, stereo pairs \cite{markovic2015sensor}, radars  \cite{mastinu2024explorations}, and depth sensors \cite{castro2022continuous}. The information from sensors enters autonomous control pipelines to enhance user performance \cite{castro2022continuous,rana2024affordance}. When using DNNs to extract relevant information, hardware accelerators equip modern robotic arms to allow on-board data processing. The resulting computing performances \cite{hundhausen2019resource, starke2022semi,chiariotti2024future} open new vistas on semi-autonomous control applications \cite{tang2022wearable,sensinger2020review}. 

 Affordance segmentation requires teleceptive sensing, to support grasping and manipulation capabilities \cite{chu2019toward}. Affordances define the possible interactions with various parts of an object, and can serve as valuable inputs for control pipelines \cite{xu2021affordance,jiang2021synergies, khalifa2022towards,wang2024sgsin}. Existing literature provides several approaches to AS \cite{hassanin2021visual, jiang2021synergies, corona2020ganhand, myers2015affordance, khalifa2022towards,ozccil2024affordance}. Hardware constraints and the presence of a user in the control loop of wearable robots call for ad-hoc solutions. The approach presented in  \cite{ragusa2021hardware} introduced a simplified, portable version of the basic AS, designed to benefit from human interaction.  The integration of DNN modules \cite{apicella2021affordance, lugani2023lightweight} and the inclusion of additional sensory inputs \cite{ragusa2023affordanceAP} further enhanced that solution. The research described in \cite{ragusa2023affordance} proved  that networks could run in real-time on high-end MCUs and support semi-autonomous control pipelines. 

Depth sensors are essential for contactless sensing \cite{krausz2019survey, chiariotti2024future}. Their integration  in deep learning pipelines has been studied  extensively \cite{deng20213d, gao2024dense, ragusa2023affordanceAP}. Conventional approaches to handling RGB-D data typically involve a two-tier architecture, in which sensor processing proceeds in separate branches and merge eventually \cite{zhou2024dgpinet}. This schema significantly increases the computational cost as compared with RGB-only pipelines. In addition, RGB-D cameras require essential calibration steps for aligning sensed data.

Hardware requirements characterize three main groups of implementation approaches to AS: unconstrained models, hardware-efficient models, and tiny models. Categories roughly identify three corresponding families of HW platforms: GPUs, application processors (embedded accelerators), and microcontrollers. Most literature focused on unconstrained models and tiny models; at the same time, mid-range, hardware-efficient solutions remain quite unexplored. To the best of the authors' knowledge, the research presented in \cite{ragusa2021hardware} only developed an AS architecture that was suitable for embedded accelerators. 
This paper shows that using embedded accelerators in the design of DNNs for RGB-D inputs can prove especially interesting for AS  in wearable robots, since one can suitably trade off accuracy and hardware requirements.   

\subsection{Hardware-aware Neural Architecture Search}

HW-NAS \cite{saha2022machine} typically drives the design of tiny, hardware-efficient DNN models, since it allows to tailor the network architectures to the target hardware platforms. When adopting HW-NAS \cite{benmeziane2021comprehensive}, one should  define a search space, a selection criteria, and a research strategy.  

The \emph{search space} covers the set of admissible candidate networks. Several studies showed that the design of the search space is crucial to the eventual performance \cite{salehin2024automl}; Popular general-purpose search spaces have been established \cite{tan2019mnasnet}. 

Defining the \emph{selection criteria} poses a critical challenge in practical implementations, due to the limitations of embedded devices \cite{zhang2020fast, li2020deep}. Aligning the optimization process with available hardware resources requires careful attention to prevent performance degradation \cite{li2021hw}. The MCUnet features a comprehensive procedure to select the architecture and configure the computing layer, yielding excellent performances \cite{lin2022ondevice}. However, the presence of a custom software layer can complicate the use and adjustment of the model \cite{banbury2021micronets}.

The main issues in adopting a \emph{research strategy} lie in its computational cost and the diversity of target platforms. Effective design approaches for tiny networks are often tailored to specific devices \cite{capogrosso2023machine}. Significant efforts have been made to accelerate the search procedure while maintaining useful constraints \cite{guo2020single, burrello2023enhancing}, but the generalization of these methods remains a topic of debate \cite{song2024efficient, li2024zero}.

\section{Designing efficient portable sensing systems based on DNNs for RGB-D data}\label{sec:proposal}

This work considers two different approaches to the design of an end-to-end pipeline, to process RGB-D inputs and feed a control logic with an affordance map. This task is supported by a small-size circuitry interfaced with a sensor, encapsulated within a portable framework. %This preprocessing  block lies within the control pipeline, and allows both to benefit from RGB-D information and take into account the  requirements of the target application and user needs.}

\begin{figure}[htbp]
    \centering
    \begin{subfigure}[b]{\linewidth}
        \centering
        \includegraphics[width=\linewidth]{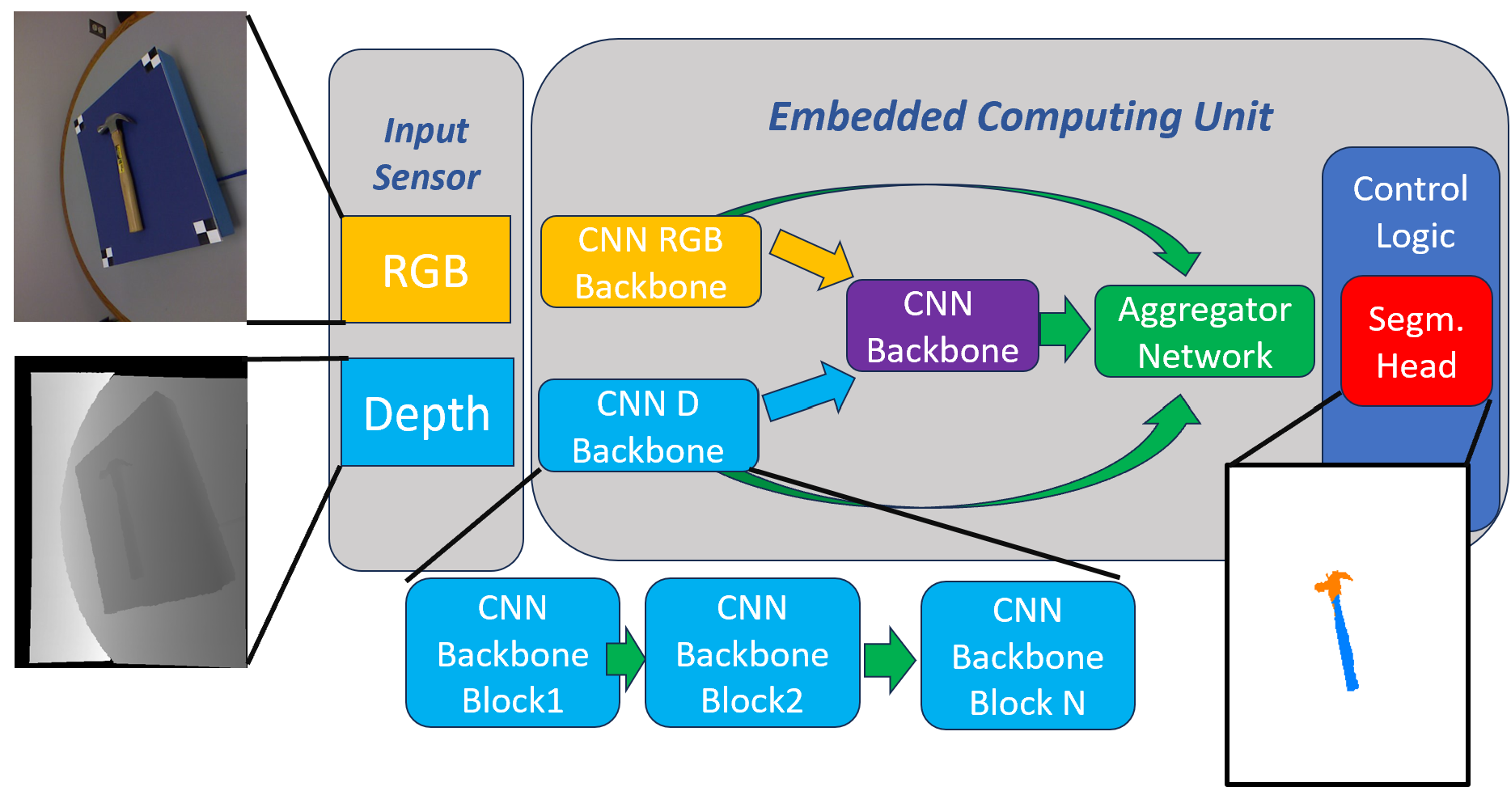}
        \caption{Proposal based on NAS}
        \label{fig:system_a}
    \end{subfigure}
    \hfill
    \\
    \begin{subfigure}[b]{\linewidth}
        \centering
        \includegraphics[width=\linewidth]{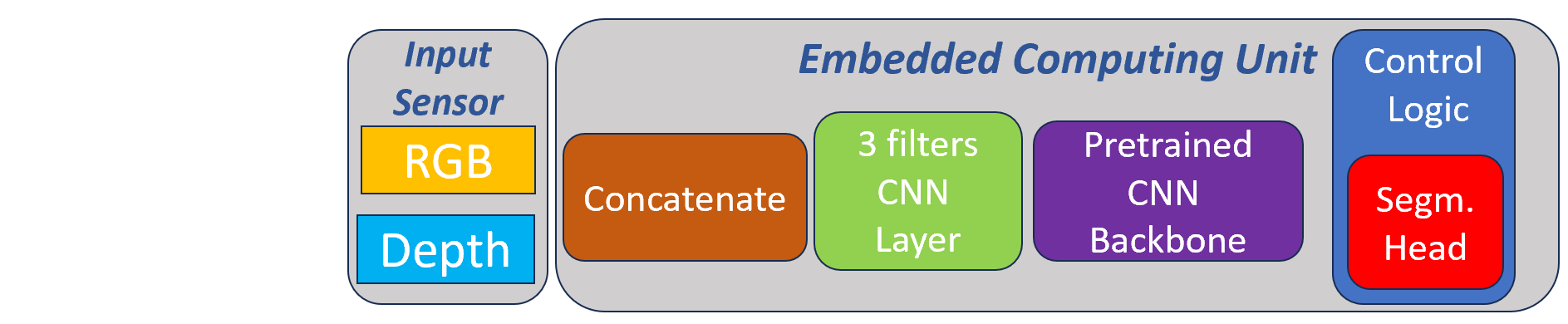}
        \caption{Proposal based on finetuning}
        \label{fig:system_b}
    \end{subfigure}
    \caption{Target sensing system per la segmentazione di affordance near-sensor usando camere RGB-D.}
    \label{fig:system}
\end{figure}

Figure \ref{fig:system}(a) outlines the first approach, which applies HW-NAS for the architecture design. Two branches process the sensor inputs (RGB abd D) separately, and merge into a CNN-backbone module that combines their results. An Aggregator network yields a unified tensor; it combines the output of the backbone with intermediate representations extracted by multiple layers of the architecture. A Segmentation head generates the final affordance mask that feeds the control logic. 
Figure \ref{fig:system}(b) illustrates the second approach. In this case, the goal is to exploit fine-tuning to include depth information on a pre-trained optimized architecture for RGB inputs.  

In both cases, the main challenge is to develop an end-to-end pipeline that a) meets accuracy requirements and b) runs on an embedded device that satisfies the constraints on energy consumption and real-time performances. The following sections will describe both approaches in detail. 

%\begin{figure}
%    \centering
%    \includegraphics[width=\linewidth]{Figures/System.png}
%    \caption{Target sensing system for near-sensor affordance segmentation using RGB-D cameras.}
%    \label{fig:system}
%\end{figure}

\subsection{HW-NAS for embedded AS}

HW-NAS provides a viable solution to the design of the DNNs as per Fig. \ref{fig:system}(a). An HW-NAS aims to find the best architecture, $a$*, given a search space $\mathcal{A}$ under a set of constraints. Two key factors drive the selection of $a$*, limiting the set of admissible networks: (1) the amount of flash and RAM available on the target device, and (2) the inference time. Empirical evidence shows that the latter factor has a crucial role because it affects the choice of admissible networks \cite{ragusa2023affordance}. Moreover, $a$* depends on the network inputs, as different data types require specific configurations, thus complicating  the selection process. The optimization problem is formalized as:

\begin{mini}|s|
{a \in \mathcal{A}} {\mathcal{L}_{val}(w^*(a),a) }
{}{}
\addConstraint{w^*(a) =argmin_w \mathcal{L}_{train}(w,a)}
\addConstraint{\phi(a)< Th}
\label{eq:optprob}
\end{mini}
where $\phi(a)$ is an estimate of the inference time on the target device and $Th$ sets the real-time constraint. 

The definition of $\mathcal{A}$ impacts on the overall performance. In the AS application (Figure \ref{fig:system}), the RGB-D input data comprise two tensor streams that, while highly correlated, convey distinct information. In practice, depth can be viewed as a grayscale image where each pixel represents a measured distance. Due to technological limitations, depth data often convey a significant amount of noise.

The RGB, D, and RGB-D backbones therefore include stacks of CNN-based encoders. CNNs are the most common choice for processing images.  The literature presents increasingly efficient implementations of vision transformers, particularly those based on custom implementations of the multi-head attention layers, which make these models well-suited for deployment on constrained devices, such as smartphone chipsets and embedded accelerators   \cite{xie2021segformer}. In this paper, the implementation relies on convolutional layers, but the overall architecture could be adapted to transformer blocks.

The base block involves a refined version of that used in MobileNetV3, without the Squeeze-and-Excite layers. Table \ref{tab:ES} summarizes the  architecture of an encoder block: each row corresponds to a layer; the first and last column represent the input and output of a layer, respectively, whereas the center column reports on the implemented operation and the associate parameters for the optimization algorithm. The search space allows to stack as many blocks as needed (Fig. \ref{fig:system_a}). This allows to  optimize the RGB and the D Backbone structures independently, as highlighted in yellow and light blue in Figure \ref{fig:system_a}. The backbone (purple block)  also includes a stack of decoder blocks, with an initial concatenation layer. 

The Aggregator block, shown in green, operates as a stack of decoder blocks, which consist of a single Conv2DTranspose layer followed by a non-linearity. The number of kernels and the shape of the activation function are the free parameters adjusted by the optimization algorithm. In the decoder, a stride value ensures that the actual size of the output mask matches its expected value. The search space allows connections from any encoder block to the Aggregator, thus enabling multiresolution feature sets (green arrows). Admissible network combinations should always include a connection to the topmost block of the RGB-D backbone.

A conventional evolutionary approach rules the optimization problem \cite{ragusa2023affordance}.  The iterative algorithm generates candidate architectures ('children') by applying random mutations to a 'parent' architecture. Children are then trained separately, and the most promising candidate is selected based on a predefined metric. This selection identifies a new parent architecture, and the process iterates until a stopping criterion is met. This straightforward approach has been adopted for its flexibility, as it avoids formal constraints on the optimization problem and the definition of the search space.

\begin{table}[]
    \caption{The Encoder block}
    \centering
    \begin{tabular}{|c|c|c|}
    \hline
   Input  & Operation & Output\\
    \hline
    Block\_In & Conv2D(exp\_f, ks = (1, 1), strides) & o1 \\
    o1  & BatchNormalization & o2 \\
    o2 & Activation & o3\\
    o3 &  DepthwiseConv2D(ks) & o4\\
    o4 & BatchNormalization & o5\\
    o5 &  Activation & 06\\
    o6 & Conv2D(squeeze, Ks = (1, 1)) & o7 \\
    o7 & BatchNormalization & o8 \\
    o8 & Conv2D(squeeze, ks = (1, 1), stride) & o9\\
    Block\_In, o9 &  Add & Block\_out\\
    \hline
    \end{tabular}
    
    \label{tab:ES}
\end{table}

\subsection{Fine tuning-based approach} \label{subsec:finetuning}
HW-NAS may exhibit some drawbacks. Firstly, it may involve remarkable computational costs for training. Secondly, it requires large labeled datasets; this can limit its applicability, since obtaining labeled masks for AS can be very time-consuming. Finally, the definition of affordance can vary depending on the context, and potentially require network retraining. Therefore, the research presented in this paper also envisions an alternative design strategy that does not involve HW-NAS. The main goal is to apply a fine-tuning process for including depth information in a pre-trained, highly optimized architecture for RGB inputs. 

As shown in Fig. \ref{fig:system}(b), the RGB and D channels join to form a tensor, $X \in R^{W \times H \times 4}$. That tensor enters a convolutional layer that includes 3 filters and a stride of 1, yielding an output tensor, $X' \in R^{W \times H \times 3}$. Thus the original architecture (optimized for RGB data) can process the latter result 'as is'. The associate fine-tuning approach brings about minimal additional costs, especially as compared with other methods.

The additional layer integrates in the architecture, and its filter values are learned by conventional backpropagation. This allows to optimize the input mapping while preserving the pre-existing weight configuration. One might argue that altering the base layer of a deep network compromises fine-tuning (retraining typically affect the topmost layers). In fact, depth maps share many structural similarities with RGB data; as a consequence, features optimized for RGB images are likely to work well for depth data, as well. This allows the optimization process to adjust low-level features and incorporate depth information, without significantly affecting the original network configuration.

\section{Experiments}\label{sec:exp}
This section is organized as follows. First, Sec. \ref{subsec:setup} describes the experimental setup. Then,  Sec. \ref{subsec:expdepth} assesses the impact of depth sensors on AS. Sec. \ref{subsec:exparch} details the structure of optimized network architectures generated by the HW-NAS to provide insights on the effective use of depth sources. Sec. \ref{subsec:expcomp} compares the proposed solutions with state-of-the-art methods and empirically proves that the approach adopted in this work can improve the trade-off between computational requirements and generalization performance. Sec. \ref{subsec:expinput} analyzes the impact of the input size on the computation cost and accuracy. Sec. \ref{subsec:expout} offers a visual analysis of the actual impact of depth sensors on affordance segmentation. Finally, Sec. \ref{subsec:expdepl} deals with the deployment of the proposed models on a prototype consisting of a Jetson Nano interfaced with a RealSense RGB-D camera.

%    \item Demonstrate that the combined use of RGB and depth data can narrow the gap between hardware-efficient approaches and large models by analyzing the Pareto optimal front for these competing objectives.

\subsection{Setup}\label{subsec:setup}
The code was developed in Python using TensorFlow and Keras libraries. Two well-known datasets, UMD \cite{myers2015affordance} and IIT \cite{nguyen2016detecting}, provided the benchmarks. The University of Maryland (UMD) dataset includes 28,843 RGB-D images across 7 object categories, with multiple framing angles for each object, allowing an assessment of the models' ability to handle framing variations. The Italian Institute of Technology (IIT) dataset includes 8,835 images under varying framing, lighting conditions, occlusion levels, and resolutions. Each dataset features distinct object sets, serving as separate benchmarks.

Validation sets were created from the training set using a standard holdout method, and the test patterns were never used for parameter or hyperparameter tuning.

Following the setup adopted by state-of-the-art works, the learning task became a three-class, pixel-wise classification problem, where the model must distinguish between 'grasp,' 'do not grasp,' and 'background'. All grasping affordances were grouped into a single class labeled 'grasp,' while all other affordances were assigned to the 'do not grasp' class \cite{ragusa2021hardware}. Foreground images of the objects were extracted using the corresponding segmentation masks, isolating the bounding boxes around the objects. Due to the different shapes of objects, the height-to-width ratios of these boxes varied. We then extracted the smallest square bounding box that fully contained each object using the AS masks. After this processing, the training and test datasets contained, respectively, 23,708 and 5,135 samples for UMD, and 9,186 and 1,969 samples for IIT. The code was executed on two workstations equipped with RTX 4080 Ti GPUs.

All parameter values were set to their default values in Keras when not specified. The NAS ran for 100 generations; the threshold for the number of FLOPS was set to the FLOPS required by MobileNetV3 in Keras to perform inference on a single image with a resolution of 128x128, i.e., 528 KFLOPS. All networks were trained for 10 epochs following the early stopping strategy described in \cite{tan2019mnasnet}. The architecture selected by NAS was re-trained for 100 epochs, with an initial learning rate of $10^{-3}$, a learning rate reduction on plateau, and early stopping based on validation loss. Generalization performance was assessed using the test set.

In the approach based on fine tuning (Sec. \ref{subsec:finetuning}), MobileNetV3 was adopted as backbone. MobileNetV3 has relatively low computational requirements while maintaining satisfactory performance on standard computer vision tasks, such as ImageNet classification. The model checkpoint, pre-trained on the ImageNet dataset, was downloaded from the Keras application repository. The same segmentation head described in \cite{lugani2023lightweight} was incorporated into the architecture. The settings used for the final architecture selected by NAS were also applied in the fine-tuning.

Network hardware requirements have been estimated using parameters and FLOPS. These measures provide an estimation independent of the target deployment devices. However, target-specific measurements, based on hardware in the loop, may exhibit different trends depending on the specific SDK.

\subsection{Efficient usage of depth information\label{subsec:expdepth}}
\begin{figure}[htp]
    \centering
    \subfloat[]{%
        \includegraphics[width=0.50\linewidth]{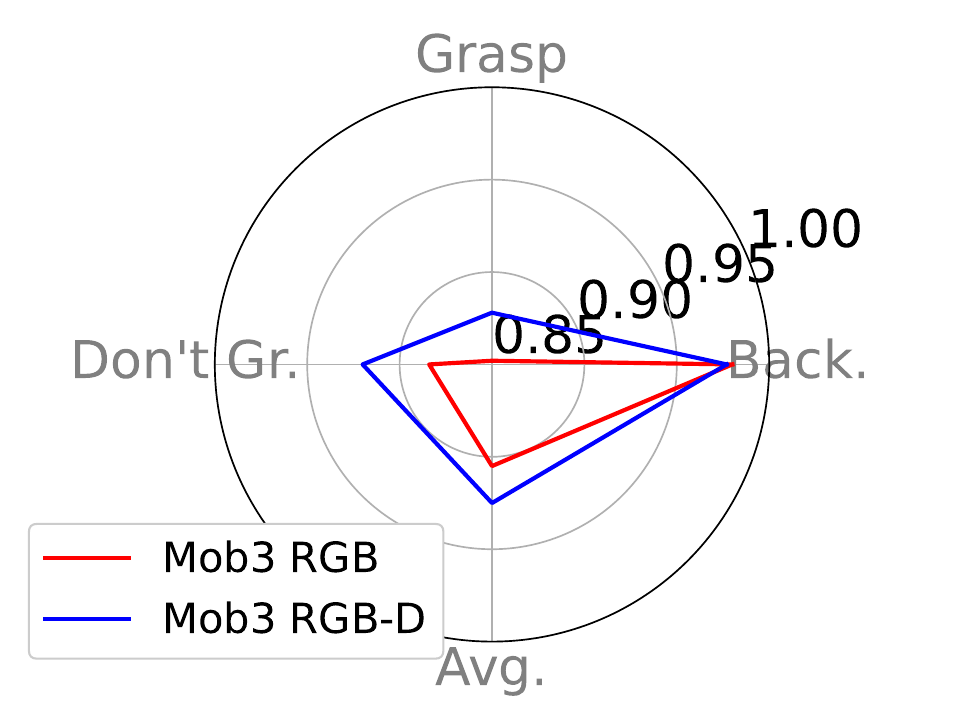}
        \label{fig:radmob}
    }
    \subfloat[]{%
        \includegraphics[width=0.50\linewidth]{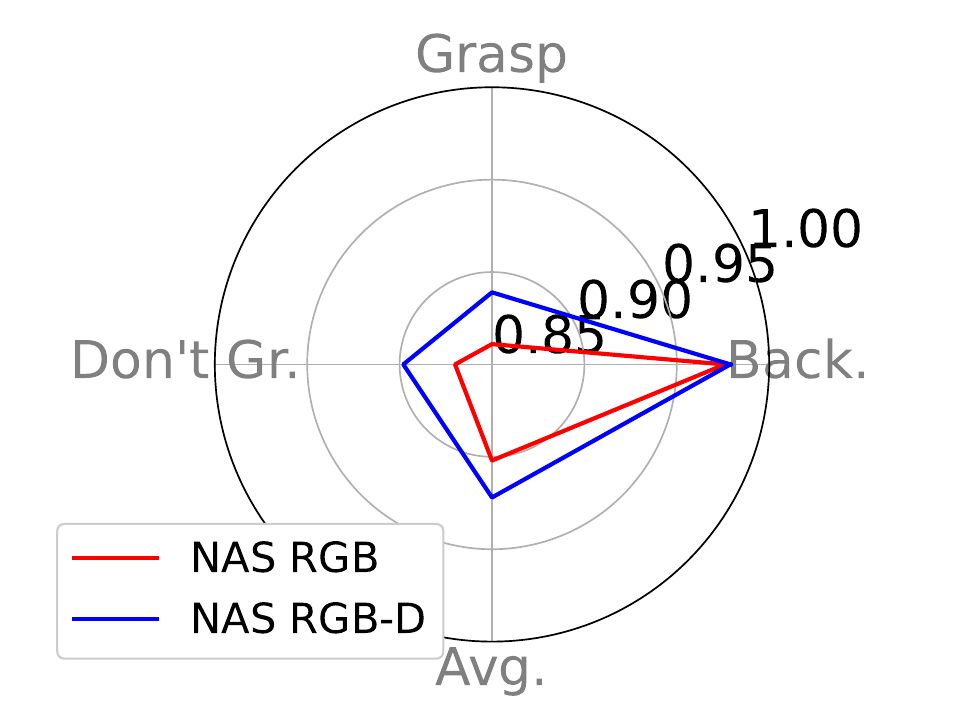}
        \label{fig:radnas}
    }
    \caption{Comparison of MobileNetV3 (a) and HW-NAS (b) performance with depth inputs for UMD.}
    \label{fig:comparison}
\end{figure}

 This section evaluates the capability of the proposed approach to efficiently utilize depth information by analyzing both generalization performance and hardware requirements.
Figure  \ref{fig:comparison} shows the generalization performance achieved with the two approaches on the UMD dataset based on the use of depth data. In both cases, the radar plot gives the accuracy obtained for the three classes and the average classification score;  the end-to-end pipeline fed with RGB-D data is compared with the  pipeline fed with RGB data. 

Figure \ref{fig:radnas} refers to the approach involving the HW-NAS. In the experiments without depth, the search space was simplified by removing the D Backbone. The availability of depth information led to an increase in accuracy across all classes. The improvements were most pronounced for the 'Grasp' (2.8\%) and 'Don't Grasp' (2.8\%) classes, which are generally more challenging.

%\begin{figure}
%    \centering
%    \includegraphics[width=0.8\linewidth]{Figures/RadarNAS}
%    \caption{Performance of HW-NAS for UMD with depth inputs.}
%    \label{fig:radnas}
%\end{figure}

Figure \ref{fig:radmob} refers to fine-tuning-based approach. Again, the network using depth data outperforms the baseline for the 'Grasp' and 'Don't Grasp' classes by 2.6\% and 3.6\%, respectively. For the 'Background' class, a negligible deterioration of 0.3\% is observed, which could be attributed to statistical fluctuations in the optimization process. Eventually, both tests confirmed the benefit of depth information in improving the model's generalization performance.

%\begin{figure}
%    \centering
%    \includegraphics[width=0.8\linewidth]{Figures/RadarMobile}
%    \caption{Performance of MobileNetV3 finetuned for UMD with depth inputs.}
%    \label{fig:radmob}
%\end{figure}

Table \ref{tab:costD}  clarifies the impact of depth processing by comparing the computational requirements of the four networks analyzed above, showing FLOPs and parameter counts for both versions. The NAS-generated networks differ both in FLOPs and parameters, as HW-NAS adjusts the architecture to maximize the impact of each layer. The number of parameters more than doubles when depth information is included, and the FLOPs increase from 257.1M to 393.2M. Nonetheless, the network requirements remain below the threshold set by the optimization procedure. Therefore, these increments should not be viewed as drawbacks of the designed architecture, which meets all specified requirements. Instead, they demonstrate that the proposed procedure can scale effectively to design larger networks when needed. The fine-tuned RGB-D model closely matches the requirements of the RGB version. The difference in the number of parameters is negligible, and the increase in FLOPs is 0.8M, i.e., the 0.1\%. This analysis confirmed that depth has a minor impact on the computing cost of the processing pipeline.

\begin{table}[]
  \caption{Summary of the computing requirements}
   \centering
   \begin{tabular}{|c|c|c|}
    \hline  
    model  & Flops & Params \\
        \hline
    NAS RGB & 257.1M & 21.6K \\
    NAS RGB-D  & 393.2M & 53.9K \\
    \hline
    Mob3 RGB & 528,5M & 918K\\
    Mob3 RGB-D & 529,3M & 918K\\

    \hline

    \end{tabular}
  
    \label{tab:costD}
\end{table}

\subsection{A detailed analysis of the generated architectures}\label{subsec:exparch}
An in-depth inspection of the architecture generated by the HW-NAS highlights the network's computational requirements. Tables \ref{tab:RGB} and \ref{tab:RGBD} provide a synopsis of, respectively, the architecture selected by the NAS for the RGB input and the architecture selected for the RGB-D input. Table \ref{tab:RGB} is divided into two parts: the first half gives details on the stack of encoders that corresponded to the backbone ; the second half details the stack of decoders that corresponded to the Aggregator. The 'Input' column identifies the name of the input fed to each block, while the 'Output' column identifies the name of the generated output. For example, in Table \ref{tab:RGB}, 'd0' is the output tensor generated by the first block, which then becomes the input to the second encoder and to two decoders. The remaining columns display the values assigned to the parameters characterizing the specific block. Skip connections play a crucial role by linking two of the three decoders to the first input block. This is consistent with some segmentation architectures, where connections to the early layers help preserve geometrical information. The proposed procedure automatically tunes these connections, advancing beyond the previous HW-NAS for AS \cite{ragusa2023affordance}, which focused on the backbone structure while relying on hand-crafted solutions for the decoder. The procedure also selected relatively large kernel sizes, up to $9×9$, which is an uncommon choice in manually designed architectures. Finally, a stride of 1 was used in most blocks, maintaining high spatial resolution, likely due to the relatively small input resolution.

\begin{table}[]
\caption{Generated RGB architecture}
    \centering
    \begin{tabular}{|ccccccc|}
    \hline
    \textbf{backbone} &  &  & & & & \\
    output & input & act. & ks & exp\_f.& squ. & stride \\ 
    \hline
    
    d0 & input &  hd\_sig. & 6 & 10 & 22 & 1\\
    
    d1 & d0 & relu & 4 & 16 & 24 & 1\\
    
    d2  & d1  & silu & 2 & 5 & 20 & 1\\
    
    d3 & d2 & relu & 8 & 26 & 21 & 2 \\
    
    d4 & d3 & relu & 7 & 28 & 21 & 1 \\
    
    d5 & d4 & relu6 & 2 & 16 & 7 & 1\\
    
    d6 & d5 & relu & 4 & 21 & 25 & 1 \\
    
    d7 & d6 & relu & 6 &  13 & 19 & 2 \\
    
    d8 & d7  & silu & 9 & 2 & 6 & 1 \\
        \hline
    
   \textbf{aggregator}&  & & & & &  \\
    output& input& act. &  n\_filt & ks &   & \\
        \hline
    h1 & d8 & relu6   & 28 & 5  & &  \\
    h2 & d0 & hd\_sig. & 19 & 2  & & \\
    h3 & d0 & hd\_sig. & 16 & 1  & &  \\
    \hline

    \end{tabular}
    
    \label{tab:RGB}
\end{table}

\begin{table}[]
   \caption{Generated RGB-D architecture}
    \centering
    \begin{tabular}{|ccccccc|}
    \hline
    \textbf{backbone} &  &  & & & & \\
    output & input & act. & ks & exp\_f.& squ. & stride \\ 
    \hline  
    \textit{RGB branch}  & & & & & &  \\
    rgb0 & in\_rgb &  relu & 5 & 26 & 28 & 1\\
    grb1 & rgb0 & silu & 5 & 21 & 25 & 1\\
    \textit{D branch}  & & & & & & \\
    d0 & in\_d  & hd\_sgm & 2 & 23 & 30 & 1 \\
   \textit{backbone}&  & & & & &  \\

    b0 & [rgb1,d0] & relu6 & 6 & 11 & 30 & 2 \\
    b1 & b0 & relu & 7 & 6 & 2 & 1\\
    b2 & b1 & silu & 5 & 2 & 21 & 1 \\
    b3& b2 & relu & 6 &  4 & 14 & 2 \\
    b4 & b3  & hd\_sgm  & 7 & 5 & 10 & 2 \\
    b5 & b4  & hd\_sgm  & 9 & 23 & 16 & 1 \\
    b5 & b4  & relu6  & 2 & 20 & 9 & 1 \\
    b6 & b5  & hd\_sgm  & 6 & 1 & 13 & 1 \\
    \hline
    \textbf{aggregator}&  & & & & &  \\
    Output& input & act. &  n\_filt & ks &   & \\
    \hline
    h1 & b7 & hd\_sig. & 24 & 9  & &  \\
    h2 & b2 & hd\_sig. & 30 & 4  & & \\
    h3 & rgb1 & relu6    & 26 & 2  & &  \\
    \hline

    \end{tabular}
 
    \label{tab:RGBD}
\end{table}

Table \ref{tab:RGBD} refers to the architecture fed with RGB-D data. Thus, it includes the RGB backbone and the D backbone. Two blocks handle the RGB input alone, while only one block processes the depth input independently. The input sources are then merged early in the computational graph, followed by a relatively long branch composed of eight blocks that process the aggregated information. This structure aligns with the fine-tuning-based approach, where RGB-D inputs are concatenated immediately. Additionally, a connection to the earliest layers is present in the architecture, underscoring the importance of skip connections for preserving geometrical information. 

\subsection{Comparison with State-of-the-Art}\label{subsec:expcomp}
Seven deep networks from the literature set the baseline for the comparison. The custom version of MobileNetV3 (CustomMobV3), proposed in \cite{ragusa2021hardware}, is an architecture designed for AS in wearable robotics. This backbone, a refined version of the standard MobileNetV3, is similar to the model adopted in this study for the fine-tuning approach, suggesting a likely overlap in results with the proposed 'Mob3 RGB' architecture.

Three lightweight models cover the area of "tiny" approaches, i.e., models optimized for the hardware constraints of microcontrollers; these architectures set a reference for computational cost, albeit with reduced generalization performance. The first model, TinySeg, is an architecture for object segmentation intended for implementation on a prosthetic hand \cite{hundhausen2019resource}. The other two tiny networks, HW-0.5 and HW-1s, were derived using the approach proposed in \cite{ragusa2023affordance} with inference time constraints of 0.5 and 1 second, respectively, on an H7 processor.

 Among the existing transformer-based architecture, SegFormer MiT-B0 \cite{xie2021segformer} has been selected as a representative example of small-size transformer that could fit constrained scenario with a reduced number of parameters and a relatively small FLOPs requirement. 

Finally, two large convolutional models serve as reference for accuracy: two networks based on EfficientNetB0 (EFF) and VGG16 (VGG16), each paired with a U-Net segmentation head \cite{li2022survey}.

\begin{figure}
    \centering
    \includegraphics[width=\linewidth]{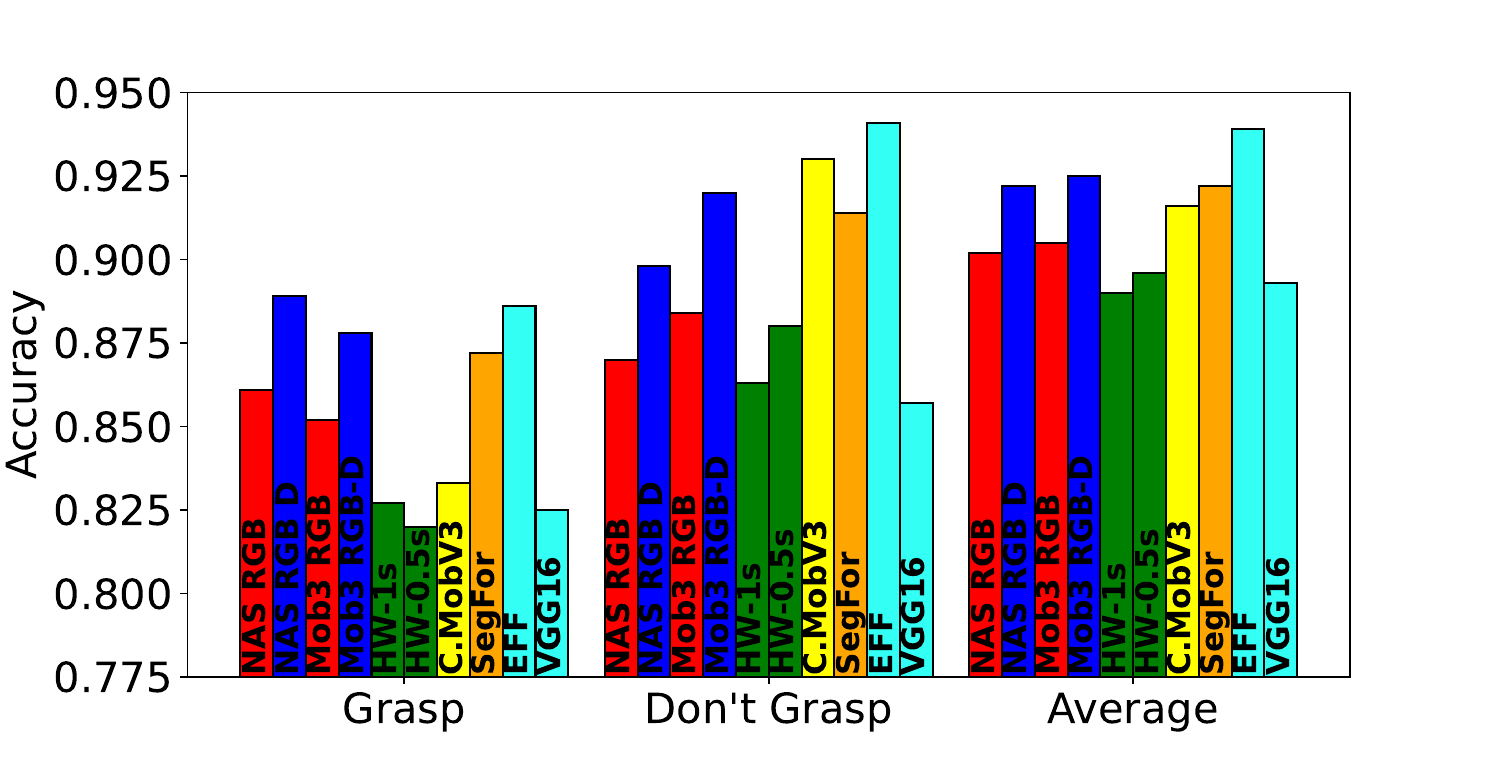}
    \caption{Comparison of the generalization performance with existing models}
    \label{fig:SOTA_GP}
\end{figure}

\begin{figure}
    \centering
    \includegraphics[width=\linewidth]{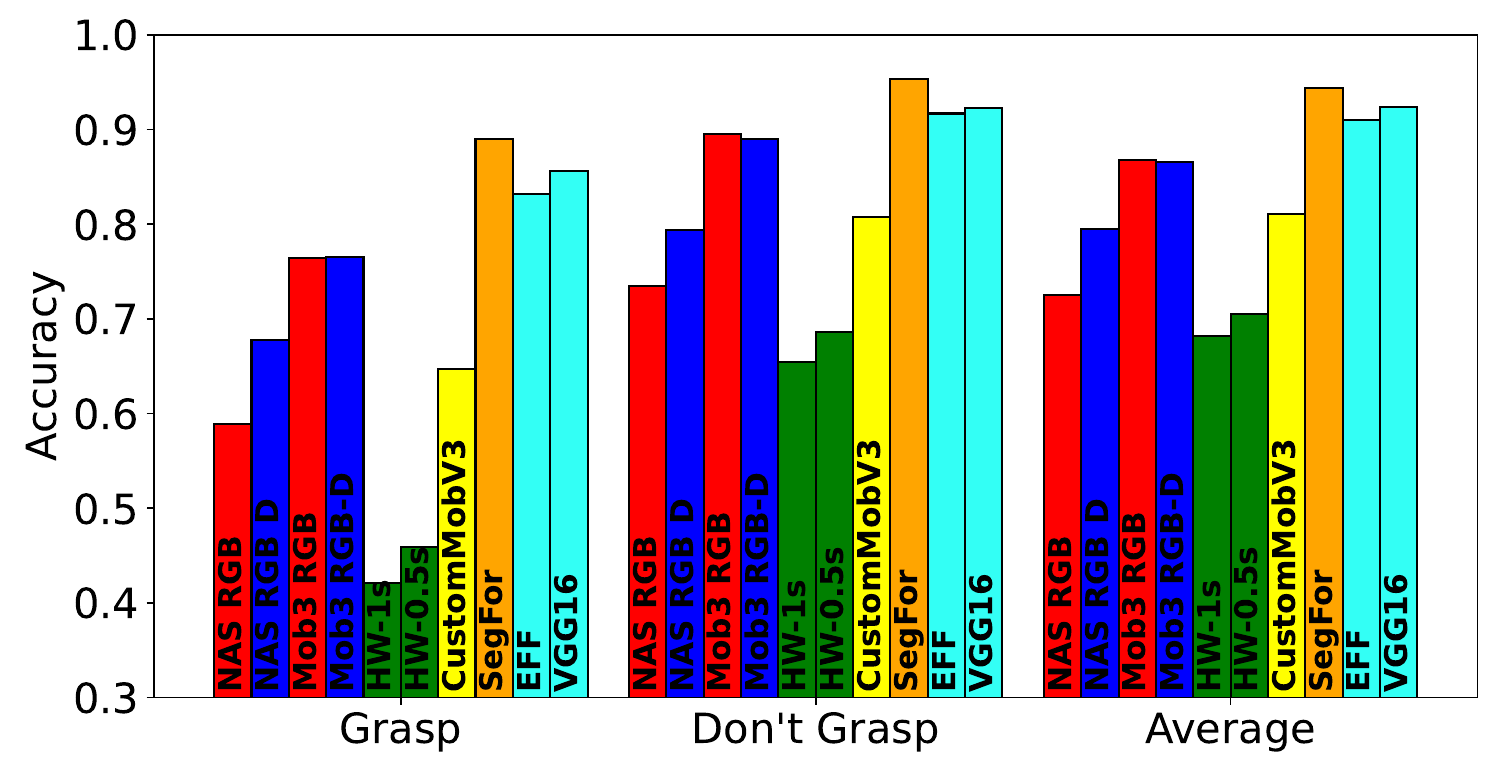}
    \caption{Comparison of the generalization performance with existing models for IIT dataset}
    \label{fig:SOTA_GP_IIT}
\end{figure}

Figure \ref{fig:SOTA_GP} compares the generalization performance on the UMD dataset of NAS-RGB, NAS-RGBD, Mob3-RGB, and Mob3-RGBD with those of the seven baselines. The plot provides the accuracy for the 'Grasp' class and for the 'Don't Grasp' class, and the weighted average class accuracy. The models are organized into  five subgroups. The first subgroup includes the proposed approaches: the model using RGB-D data is marked in blue, while the RGB-only counterpart is marked in red. Tiny models are shown in green. The portable model based on the custom MobileNetV3 (CustomMobV3) is represented in yellow. Small-scale transformer models are represented by SegFormer MiT-B0 in orange while large convolutional models are shown in light blue. Figure \ref{fig:SOTA_GP_IIT} compares the generalization performance on the IIT dataset, which introduces greater variability in background, illumination, and occlusions. This figure and Fig. \ref{fig:SOTA_GP} share the same format. Note that both plots do not give the performance of TinySeg, one of the tiny models included in the comparison. In fact, the accuracy of TinySeg was below 77.5\% in the case of Fig. \ref{fig:SOTA_GP} and below 30\% in the case of Fig. \ref{fig:SOTA_GP_IIT}.    

The proposed models with RGB-D input always outperformed tiny models (green bars in the plots). Although large models and transformers (light blue and orange bars in the plots) most of the time achieved the highest accuracy, the use of depth information significantly improved the performance of the proposed models in the case of experiments involving the UMD dataset. A slightly different trend characterizes experiments involving the more challenging IIT benchmark: the fine-tuning approach is always better than the NAS-based approach, and depth information seems to not boost accuracy.  One possible explanation is that the IIT dataset needs a more robust feature set because it involves large variations in framing quality, illumination, and background, which makes the problem more challenging. The large-size CNNs, transformer, and fine-tuned models were trained starting from checkpoints pre-trained on the ImageNet dataset. As a result, these networks leverage robust feature sets optimized on millions of images. The larger number of parameters also helps to explain the gap between the NAS-generated network and fine-tuned models.  Indeed, the transformer model (represented in orange) and the large-size CNNs outperform all the other solutions on the IIT dataset. The role of fine-tuning can be further highlighted by comparing the performance of Mob3RGB and C.Mob3 across the two datasets. In the case of the IIT dataset, the fine-tuned Mob3RGB always scored a better accuracy. The custom MobileNetV3 (yellow bar in the plots) actually topped all the proposed models only once, i.e., in the 'don't grasp' class on the UMD dataset. In general, models with RGB-D input can achieve at least the same accuracy as the custom MobileNetV3.

%the gap is reduced.  is comparable to the RGB version of the proposed approach: it performs better on the 'Don't Grasp' class but shows lower accuracy for 'Grasp,' resulting in a slight decrease in the overall average accuracy. The gap between large-size models and hardware-efficient solutions is larger for the IIT dataset compared to UMD. However, the proposed approach, which incorporates depth information, reduces this gap. In the case of NAS, the inclusion of depth input improves performance for all classes. Conversely, for MobileNetV3, the difference becomes negligible. 

\begin{figure}
    \centering
    \includegraphics[width=\linewidth]{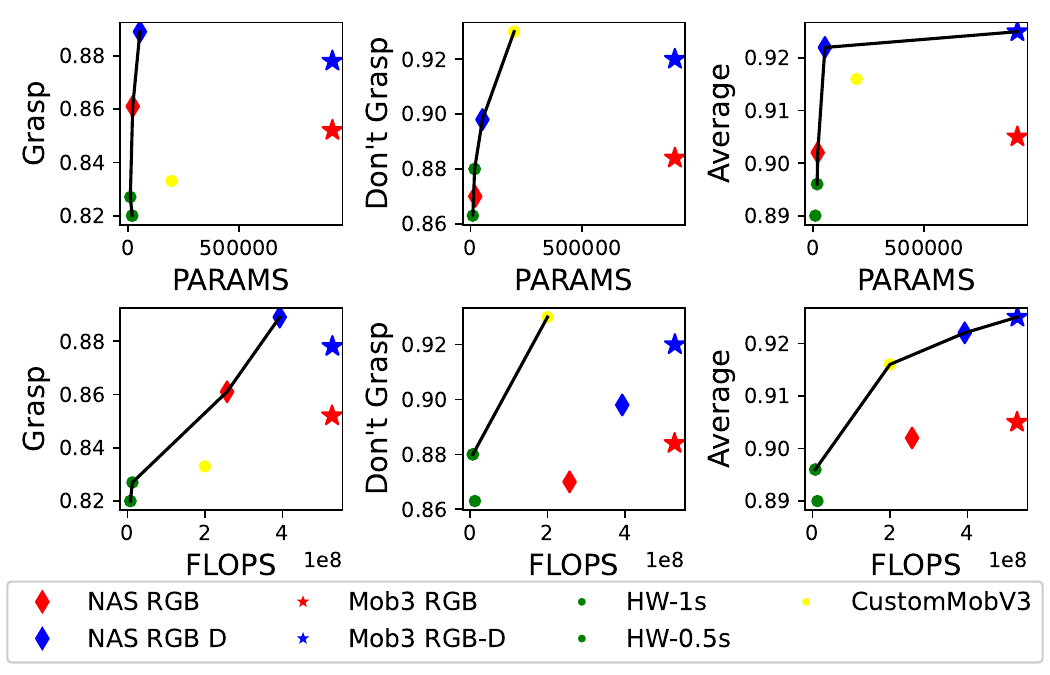}
    \caption{Analysis of the trade-off between generalization performance and resources' requirements.}
    \label{fig:GP_HW}
\end{figure}

\begin{figure}
    \centering
    \includegraphics[width=\linewidth]{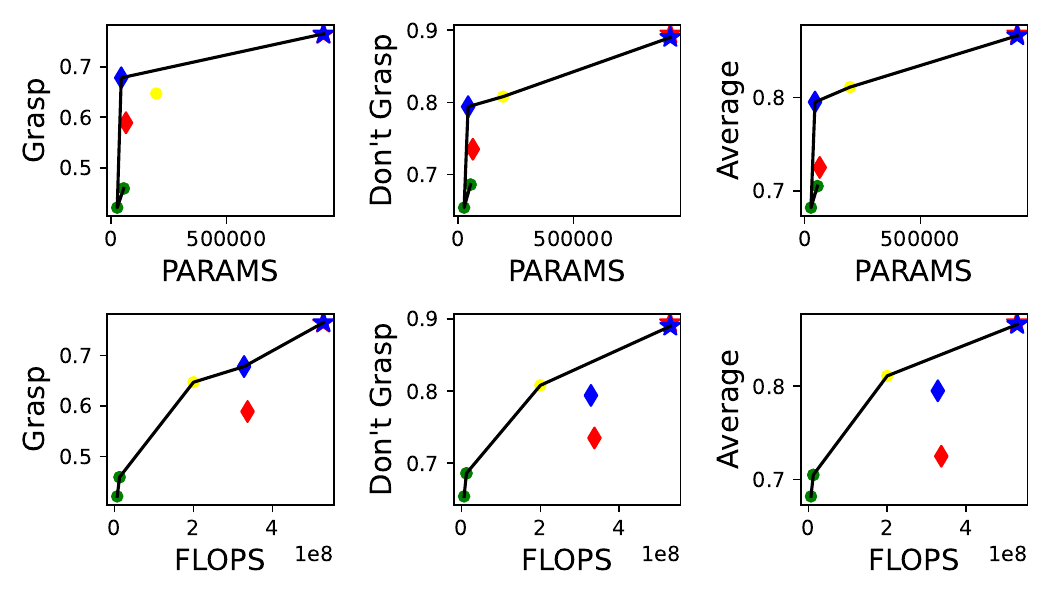}
    \caption{Analysis of the trade-off between generalization performance and resources requirements for IIT dataset.}
    \label{fig:GP_HW_IIT}
\end{figure}

Figures \ref{fig:GP_HW} and \ref{fig:GP_HW_IIT} further inspect the outcomes of the comparison taking into account the footprint of the models in terms of FLOPs and the number of parameters. 
Figure \ref{fig:GP_HW} displays six scatter plots arranged in a grid. The three columns correspond, respectively, to the comparison on the 'Grasp' class, the comparison on the 'Don't Grasp' class, and the comparison on the 'Average' performance. In each scatter plot, the $y$-axis gives the pixel-wise accuracy for the corresponding column. For instance, in the scatter plots in the central column, the $y$-axis shows the pixel-wise accuracy for the 'Don't Grasp' class. In the first row of the grid, number of parameters is on the $x$-axis; while in the second row FLOPs is on the $x$-axis. The color scheme replicates the one adopted above; thus, for example, the blue markers refer to the proposed models with RGB-D input and the red markers refer to the proposed models with RGB input. Furthermore, the diamond marker identifies, among the proposed models, those obtained from NAS, while the star marker identifies the models obtained with fine-tuning. 
Large models (i.e., the large-size CNNs and the transformer) were excluded from these plots due to their high hardware requirements, as the goal was to compare small-size models. In every plot, a Pareto optimal front highlights models that best balance accuracy and computational requirements. The same format has been adopted for Fig. \ref{fig:GP_HW_IIT}. 

In the experiments involving the UMD dataset, both the proposed models with RGB-D inputs (blue markers) remain on the Pareto optimal in the final column, which refers to the average classwise accuracy. For the 'Grasp' class, only the NAS-generated models (diamond markers) are on the Pareto front. In fact, without the NAS-based approach, the Mob3 RGB-D architecture would have been on the optimal front. The only case where the proposed models do not lie on the Pareto front is for the 'Don't Grasp' class, where the CustomMobileNetV3 (yellow marker) achieves superior performance. 
The proposed models with RGB-D input consistently lie on the Pareto optimal front -or very close to- in the experiments involving the IIT dataset (Fig. \ref{fig:GP_HW_IIT}). This in turn proves the capability of the proposed approach to a) fully exploit RGB-D input and b) balance hardware requirements and generalization performance in complex benchmarks as well.

\subsection{Impact of input size on performance}\label{subsec:expinput}
FLOPS remain the best device-agnostic estimator for inference time. %, although not directly equivalent to inference time due to the low-level characteristics of the device performing inference. 
The reliability of this estimator increases as the number of cores in the computing unit decreases. For example, in microcontrollers, the correlation between FLOPS and inference time approaches 1 \cite{ragusa2023affordance}. 

\begin{figure}
    \centering
    \includegraphics[width=\linewidth]{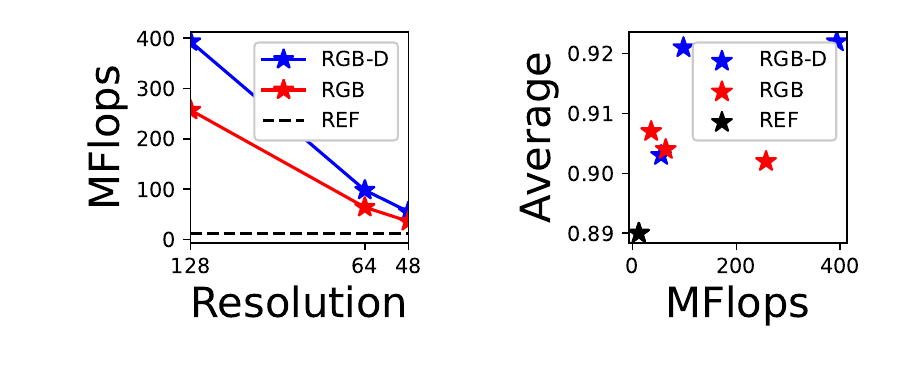}
    \caption{Left: Analysis of Flops for various Input sizes; Right: Analysis of average accuracy and Flops requirements for variable input resolutions}
    \label{fig:FlopsResTradeoff}
\end{figure}

The most straightforward but effective method of lowering  FLOPS is to reduce input resolution \cite{ragusa2023affordance}, even if this introduces practical downsides in the framing. Figure \ref{fig:FlopsResTradeoff} reports the results of an experiment in which three resolutions were considered: $128x128$ pixels, $64x64$ pixels, and $48x48$ pixels. A resolution of $128x128$ is the standard setup adopted in previous works, while $48x48$ is the resolution used in \cite{ragusa2023affordance} to meet the hard constraints of microcontrollers.

On the left side of Fig. \ref{fig:FlopsResTradeoff}, the plot gives the number of MFLOPS for a single inference as a function of the input resolution for NAS RGB-D (in blue) and NAS RGB (in red). The black dashed line is the reference set by the 12.46 MFLOPS required by HW-1S  for \(48 \times 48\) inputs \cite{ragusa2023affordance}. As the input resolution decreases, both proposed architectures approach the FLOPS of the reference. On the right side of Fig. \ref{fig:FlopsResTradeoff}, a scatter plot displays the three NAS RGB-D models and the three NAS RGB models in a chart with MFLOPS on the $x$-axis and average accuracy on the $y$-axis. This plot shows that with NAS RGB-D one can obtain an average accuracy above 0.9 even with tight constraints on the MFLOPS. Indeed, an higher average accuracy can be reached when one is allowed to increase the MFLOPS. Conversely, with NAS RGB the average accuracy seems almost independent of the FLOPS. Thus, the approach based on NAS fully exploits the information conveyed by the depth channel.

%\subsection{Occlusions and illumination impact}
%The comparison with SOTA models was repeated on the IIT dataset, which introduces greater variability in background, illumination, and occlusions. Figures \ref{fig:SOTA_GP_IIT} and \ref{fig:GP_HW_IIT} replicate the settings of Figures \ref{fig:SOTA_GP} and \ref{fig:GP_HW}.

%The gap between large-size models and hardware-efficient solutions is larger for the IIT dataset compared to UMD. However, the proposed approach, which incorporates depth information, reduces this gap. In the case of NAS, the inclusion of depth input improves performance for all classes. Conversely, for MobileNetV3, the difference becomes negligible. For this more challenging benchmark, the fine-tuning approach is always beneficial, and the addition of depth does not affect accuracy. One possible explanation is the need for more robust feature sets, which have been optimized during the original training on the larger ImageNet dataset.

\subsection{Output analysis} \label{subsec:expout}
\begin{figure}
    \centering
    \includegraphics[width=\linewidth]{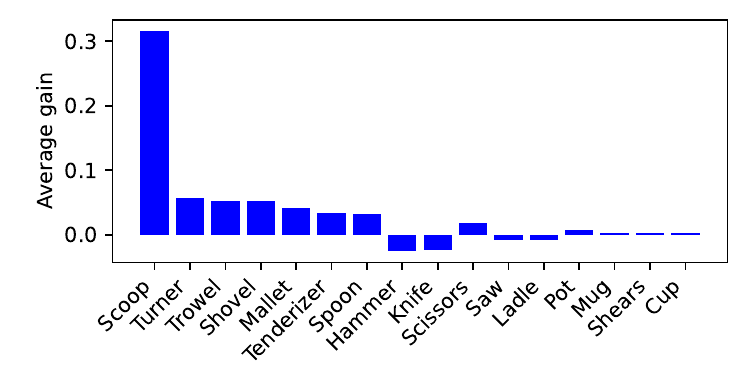}
    \caption{Gain for images of single testing objects.}
    \label{fig:ClassGain}
\end{figure}

The actual impact of depth sensors on affordance segmentation may vary significantly across different objects. Figure \ref{fig:ClassGain} shows -for each object included the UMD test set- the difference in average classification accuracy obtained by adding depth information. A positive value indicates that the model with RGB-D input achieved a greater accuracy than the model with RGB input.

\begin{figure}
    \centering
    \includegraphics[width=\linewidth]{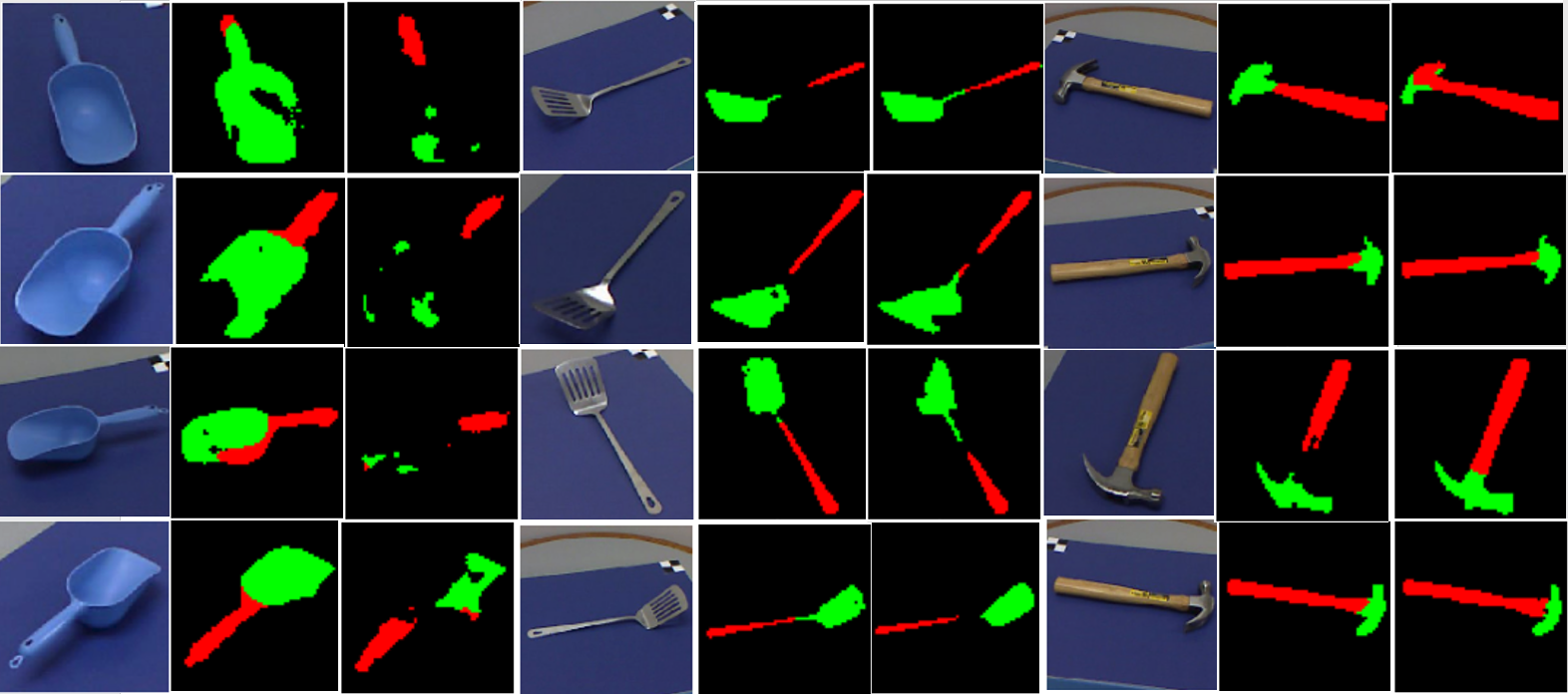}
    \caption{Examples of the impact of depth in the final prediction.}
    \label{fig:examples}
\end{figure}

In 12 cases out of 16, the availability depth information led to greater accuracy, with a difference exceeding 3\% in 7 cases. The improvement was substantial for the scoop; in this case the difference was 31.6\%. Figure \ref{fig:examples} shows examples of images taken from the dataset for 'scoop','turner', and 'hammer.' For each object, four different input images are provided along with the corresponding RGB-D segmentation mask and RGB segmentation mask. Depth information can be crucial when lighting and color conditions are suboptimal. The scoop is blue, a color very similar to the foreground used in the experimental setup. Thus, the availability of a depth sensor is very important. Even in the least favorable configurations, the decrease in accuracy was limited, highlighting the overall beneficial impact of including depth sensors in the sensing pipeline. Notably, when object color closely matched the background or when metallic surfaces caused reflection issues, depth information significantly enhanced the models' segmentation capabilities. Interestingly, even for some flat objects, depth proved beneficial, challenging the assumption of a direct relationship between object shape and the effectiveness of depth sensors.

A further analysis was conducted on a test set generated with Unreal Engine 5. This setup enabled the creation of many heterogeneous images through a cost-effective procedure, where test patterns composed of RGB-D images and affordance masks were generated using the simulator. A test set of 2,000 images was created, with objects belonging to the same classes available in the UMD dataset. Models trained on UMD were then tested on this test set, showing an improvement of 11.5\% in the average accuracy when the D information was used. This gap was reduced to 4.6\% when data augmentation was applied to mitigate cross-domain issues commonly associated with simulator-generated images. The results confirmed the advantage of using depth data to generalize to unseen test objects, thanks to the geometric information it provides.

\subsection{Deployment on embedded accelerators}\label{subsec:expdepl}
The two proposed models using RGB-D data have been deployed on a prototype consisting of a Jetson Nano interfaced with a RealSense RGB-D camera, i.e., one of the most widely used RGB-D cameras. The Jetson Nano was chosen among the available embedded accelerators because of its excellent balance between performance and power consumption. The market indeed offers newer embedded accelerators that provide an excellent trade-off between energy and performance; however, these devices are typically designed for larger models, making them less effective in this specific scenario. 

The comparison involved four models: the plain TensorFlow implementation of NAS RGBD and Mob3 RGBD; the optimized versions (\_opt) of NAS RGBD and Mob3 RGBD obtained using the TRT optimizer with float16 quantization. 

All the data on latency and memory usage provided in the following have been obtained by completing 100 inferences on RGB-D images captured by the camera. Unless explicitly stated otherwise, the reported metrics represent average values. %Furthermore, standard deviation values were omitted when they were not significant.
Memory usage was monitored by relying on the operating system's utilities. Power consumption was measured by recording the current drawn at the board's power input using a wattmeter. %In all measurements, the development board was connected to a 220V power outlet via a transformer. 
Framing conditions were not particularly relevant to the hardware requirements, as they are independent of the input image.

 \subsubsection{Frame rate}
The first analysis evaluates the inference time of the models for two power configurations of the Jetson Nano: 5W and 10W. In both cases, the value corresponds to the estimated maximum peak power consumption. Table \ref{tab:framerate} gives, for each tested model, the time needed to process a frame and the frame rate (expressed in frames per second).   

The table shows that optimization had a significant impact on the inference time for both models. Both optimized versions enable inference under 120 ms, supporting relatively fast processing suitable for real-time operations intended for human users.

\begin{table}[]
   \caption{Measured frame rate}
    \centering
    \begin{tabular}{|ccc|cc|}
    \hline
 Model & 5W & &  10W & \\ & Sec. per frame  & FPS & Sec. per frame & FPS\\
       
       \hline
MOB3 & 0.55 s & 2 & 0.39 s & 2 - 3\\
MOB3\_opt & 0.12 s & 8 - 9 & 0.07 s & 14 - 15\\
NAS & 0.36 s &3 &0.24 s &4\\
NAS\_opt & 0.11 s &  9 & 0.06 s& 15- 16\\
\hline
    \end{tabular}
 
    \label{tab:framerate}
\end{table}

\subsubsection{Memory analysis}
The Jetson Nano employs a shared memory architecture, where both processor and GPU utilize the same memory, making data transfers more efficient. Memory profiling is a non-trivial task due to the interaction between Python and the operating system. We propose three sets of measurements to analyze memory usage. 
As empirical evidence suggests that RAM usage is not affected by the power configuration, this parameter will be omitted in the following tables.

Table \ref{tab:memall} reports on the memory profiling performed in the experiments. The table is divided into three sections. Each section features two columns: the average memory usage and the peak value, which refers to the maximum memory allocated during the entire process. The red section presents the total memory usage reported by Tracemalloc for each model, including library imports and utilities. The Tracemalloc software was adopted to track memory allocated directly by Python. However, this software may not track external allocations. The measurements confirmed the impact of the architecture on memory requirements. On average, the NAS architecture uses about 300 MB of memory, while MOB3 requires 30\% more memory. It is worth noting that the optimized versions use a slightly larger amount of memory because of the use of the TRT engine. Peak memory usage can reach up to twice the average memory requirement.

\begin{table}[]
    \caption{Memory Measurements}
    \centering
    \begin{tabular}{|c|cc|cc|c|}
    \hline
    & \multicolumn{2}{c|}{\cellcolor{red!25}  Full Python} &\multicolumn{2}{c|}{\cellcolor{green!25}  Model} &\cellcolor{blue!25}  Overall \\
    \hline
Model &\cellcolor{red!25}\cellcolor{red!25}Avg.  & \cellcolor{red!25}Peak  &\cellcolor{green!25}\cellcolor{green!25}Avg.  & \cellcolor{green!25}Peak &\cellcolor{blue!25}Avg. \\
\hline
MOB3 & 408 MB & 925 MB & 283 MB & 800 MB & 3.5 GB\\
MOB3\_opt & 436 MB & 1030 MB &  311 MB & 905 MB & 4.8 GB\\
NAS & 289 MB & 587 MB & 164 MB & 462 MB & 2.6 GB\\
NAS\_opt & 292 MB & 602 MB & 167 MB & 477 MB & 3.2 GB\\
\hline
    \end{tabular}

    \label{tab:memall}
\end{table}

%\begin{table}[]
%    \centering
%    \begin{tabular}{|ccc|}
%    \hline
%Model &Average Memory & Peak Memory\\
%\hline
%MOB3 & 408 MB & 925 MB\\
%MOB3\_opt & 436 MB & 1030 MB\\
%NAS & 289 MB & 587 MB\\
%NAS\_opt & 292 MB & 602 MB\\
%\hline
%    \end{tabular}
%    \caption{Full Python's memory requirements}
%    \label{tab:pymem}
%\end{table}

The green section of table \ref{tab:memall} shows the amount of RAM required to load the model parameters and perform the inference. In this case, the trends are similar to those assessed with Tracemalloc, with changes in the absolute values.

%\begin{table}[]
%    \centering
%    \begin{tabular}{|ccc|}
%    \hline
%Model &Average Memory & Peak Memory\\
%\hline
%MOB3 & 283 MB & 800 MB \\
%MOB3\_opt &  311 MB & 905 MB\\
%NAS & 164 MB & 462 MB\\
%NAS\_opt &167 MB &477 MB\\
%\hline
%    \end{tabular}
%    \caption{Memory requirements for model loading and inference}
%    \label{tab:infmem}
%\end{table}

A third measurement was performed using the system monitor utility available on the Nano's operating system. In this case, only average values are reported in the blue section of Table \ref{tab:memall}. As expected, the total RAM usage is significantly higher. In fact, in all cases, more than 2.5 GB were used by the combination of the operating system and other processes, confirming the large number of background processes running on the OS. Nevertheless, again the architecture generated with NAS is the one that uses fewer memory.

%\begin{table}[]
%    \centering
%    \begin{tabular}{|cc|}
%    \hline
%    Model & Memory\\
%    \hline
%    MOB3 & 3.5 GB\\
%    MOB3\_opt & 4.8 GB\\
%    NAS & 2.6 GB\\
%    NAS\_opt & 3.2 GB\\
%    \hline
%    \end{tabular}
%    \caption{Overall RAM consumption of the system}
%    \label{tab:sysmem}
%\end{table}

\subsubsection{Power consumption}
A power meter with a relative error of 2\% was connected to the development board's power supply. The measurements were taken with the RealSense camera connected and powered directly by the Jetson during the acquisition and inference procedures.

Table \ref{tab:power} reports the average power consumption of the system for each model. The values have been rounded to the nearest integer to take into account the measurement uncertainty. In both configurations, the average power requirement exceeded the threshold value because the power meter measures the consumption of both the Jetson and the sensor. However, in all cases, the power consumption remains below 11 W, making the system suitable for use with portable lithium batteries.

As a simple estimation, considering an average smartphone battery with a capacity of 5000 mAh, configurations with a power consumption of 7 Watts would provide approximately 3.5 hours of operation. This is a sufficiently long lifespan, especially considering that inferences are not expected to run continuously but only upon user request.

%\begin{table}[]
%    \centering
%    \begin{tabular}{|ccc|}
%    \hline
%         Model &5W Mode & 10W Mode\\
%         \hline
%MOB3 & 7 W & 8 W\\
%MOB3\_opt & 7 W & 9 W\\
%NAS\ & 7 W  & 8 W\\
%NAS\_opt & 8 W & 11 W\\
%\hline
%    \end{tabular}
%    \caption{Average power consumption during acquisition and inference procedure.}
%    \label{tab:power}
%\end{table}

\begin{table}[]
    \caption{Average power consumption during acquisition and inference procedure.}
    \centering
    \begin{tabular}{|ccc|}
    \hline
         Model &\multicolumn{1}{c}{5W Mode (opt)} & \multicolumn{1}{c|}{10W Mode (opt)}\\
         \hline
MOB3 & 7W (7W)& 8W (9W)\\
%MOB3\_opt & & W\\
NAS\ & 7 W (8W) & 8 W(11W)\\
%NAS\_opt & 8 W & 11 W\\
\hline
    \end{tabular}

    \label{tab:power}
\end{table}

\section{Conclusion}
The paper analyzed two different design strategies for the deployment of affordance segmentation in wearable sensing systems. The results showed that by including RGB-D sensors in the sensing pipeline one can improve the generalization performance with respect to the setup involving only RGB cameras. In addition, extensive experimental analysis on two well-known real-world benchmarks proved that the proposed strategies can generate architectures lying on the Pareto-optimal front for generalization performance and hardware requirements. The architecture generated with the proposed strategies can process images in real-time with acceptable power consumption when running on a prototype based on a Jetson Nano and a Real Sense RGB-D camera. 

The current version of the system has been evaluated using foreground images, leaving the correct framing to the end users. This could indeed prove challenging. Future versions will incorporate object localization capabilities, enabling a simpler framing setup. Additionally, in future experiments, we plan to apply the same techniques to transformer models by designing dedicated search spaces or adapting the fine-tuning approach for models with positional encoding. Eventually, clinical trials will also be required to confirm the user's perceived benefits in the semi-autonomous pipeline. In this setup, the proposed system will be included in the full control pipeline, allowing an in-depth analysis of how an efficient AS can improve user interaction.  

\bibliographystyle{IEEEtran}
\bibliography{refs}

@article{markovic2015sensor,
  title={Sensor fusion and computer vision for context-aware control of a multi degree-of-freedom prosthesis},
  author={Markovic, Marko and Dosen, Strahinja and Popovic, Dejan and Graimann, Bernhard and Farina, Dario},
  journal={Journal of neural engineering},
  volume={12},
  number={6},
  pages={066022},
  year={2015},
  publisher={IOP Publishing}
}

@inproceedings{nguyen2016detecting,
  title={Detecting object affordances with convolutional neural networks},
  author={Nguyen, Anh and Kanoulas, Dimitrios and Caldwell, Darwin G and Tsagarakis, Nikos G},
  booktitle={2016 IEEE/RSJ International Conference on Intelligent Robots and Systems (IROS)},
  pages={2765--2770},
  year={2016},
  organization={IEEE}
}

@inproceedings{myers2015affordance,
  title={Affordance detection of tool parts from geometric features},
  author={Myers, Austin and Teo, Ching L and Ferm{\"u}ller, Cornelia and Aloimonos, Yiannis},
  booktitle={2015 IEEE International Conference on Robotics and Automation (ICRA)},
  pages={1374--1381},
  year={2015},
  organization={IEEE}
}

@inproceedings{hundhausen2019resource,
  title={Resource-Aware Object Classification and Segmentation for Semi-Autonomous Grasping with Prosthetic Hands},
  author={Hundhausen, Felix and Megerle, Denis and Asfour, Tamim},
  booktitle={2019 IEEE-RAS 19th International Conference on Humanoid Robots (Humanoids)},
  pages={215--221},
  year={2019},
  organization={IEEE}
}

@inproceedings{tan2019mnasnet,
  title={Mnasnet: Platform-aware neural architecture search for mobile},
  author={Tan, Mingxing and Chen, Bo and Pang, Ruoming and Vasudevan, Vijay and Sandler, Mark and Howard, Andrew and Le, Quoc V},
  booktitle={Proceedings of the IEEE Conference on Computer Vision and Pattern Recognition},
  pages={2820--2828},
  year={2019}
}

@article{sensinger2020review,
  title={A Review of Sensory Feedback in Upper-Limb Prostheses From the Perspective of Human Motor Control},
  author={Sensinger, Jonathon W and Dosen, Strahinja},
  journal={Frontiers in Neuroscience},
  volume={14},
  year={2020},
  publisher={Frontiers Media SA}
}

@article{hassanin2021visual,
  title={Visual affordance and function understanding: A survey},
  author={Hassanin, Mohammed and Khan, Salman and Tahtali, Murat},
  journal={ACM Computing Surveys (CSUR)},
  volume={54},
  number={3},
  pages={1--35},
  year={2021},
  publisher={ACM New York, NY, USA}
}

@article{weiner2022designing,
  title={Designing Prosthetic Hands With Embodied Intelligence: The KIT Prosthetic Hands},
  author={Weiner, Pascal and Starke, Julia and Rader, Samuel and Hundhausen, Felix and Asfour, Tamim},
  journal={Frontiers in Neurorobotics},
  volume={16},
  year={2022},
  publisher={Frontiers Media SA}
}

@article{khalifa2022towards,
  title={Towards Visual Affordance Learning: A Benchmark for Affordance Segmentation and Recognition},
  author={Khalifa, Zeyad Osama and Shah, Syed Afaq Ali},
  journal={arXiv preprint arXiv:2203.14092},
  year={2022}
}

@article{ragusa2021hardware,
  title={Hardware-aware affordance detection for application in portable embedded systems},
  author={Ragusa, Edoardo and Gianoglio, Christian and Dosen, Strahinja and Gastaldo, Paolo},
  journal={IEEE Access},
  volume={9},
  pages={123178--123193},
  year={2021},
  publisher={IEEE}
}

@inproceedings{apicella2021affordance,
  title={An Affordance Detection Pipeline for Resource-Constrained Devices},
  author={Apicella, Tommaso and Cavallaro, Andrea and Berta, Riccardo and Gastaldo, Paolo and Bellotti, Francesco and Ragusa, Edoardo},
  booktitle={2021 28th IEEE International Conference on Electronics, Circuits, and Systems (ICECS)},
  pages={1--6},
  organization={IEEE}
}

@article{li2022survey,
  title={A Survey of Multifingered Robotic Manipulation: Biological Results, Structural Evolvements and Learning Methods},
  author={Li, Yinlin and Wang, Peng and Li, Rui and Tao, Mo and Liu, Zhiyong and Qiao, Hong},
  journal={Frontiers in Neurorobotics},
  pages={53},
  year={2022},
  publisher={Frontiers}
}

@inproceedings{corona2020ganhand,
  title={Ganhand: Predicting human grasp affordances in multi-object scenes},
  author={Corona, Enric and Pumarola, Albert and Alenya, Guillem and Moreno-Noguer, Francesc and Rogez, Gr{\'e}gory},
  booktitle={Proceedings of the IEEE/CVF conference on computer vision and pattern recognition},
  pages={5031--5041},
  year={2020}
}

@article{benmeziane2021comprehensive,
  title={A comprehensive survey on hardware-aware neural architecture search},
  author={Benmeziane, Hadjer and Maghraoui, Kaoutar El and Ouarnoughi, Hamza and Niar, Smail and Wistuba, Martin and Wang, Naigang},
  journal={arXiv preprint arXiv:2101.09336},
  year={2021}
}

@article{li2020deep,
  title={The deep learning compiler: A comprehensive survey},
  author={Li, Mingzhen and Liu, Yi and Liu, Xiaoyan and Sun, Qingxiao and You, Xin and Yang, Hailong and Luan, Zhongzhi and Gan, Lin and Yang, Guangwen and Qian, Depei},
  journal={IEEE Transactions on Parallel and Distributed Systems},
  volume={32},
  number={3},
  pages={708--727},
  year={2020},
  publisher={IEEE}
}

@inproceedings{zhang2020fast,
  title={Fast hardware-aware neural architecture search},
  author={Zhang, Li Lyna and Yang, Yuqing and Jiang, Yuhang and Zhu, Wenwu and Liu, Yunxin},
  booktitle={Proceedings of the IEEE/CVF Conference on Computer Vision and Pattern Recognition Workshops},
  pages={692--693},
  year={2020}
}

@article{li2021hw,
  title={Hw-nas-bench: Hardware-aware neural architecture search benchmark},
  author={Li, Chaojian and Yu, Zhongzhi and Fu, Yonggan and Zhang, Yongan and Zhao, Yang and You, Haoran and Yu, Qixuan and Wang, Yue and Lin, Yingyan},
  journal={arXiv preprint arXiv:2103.10584},
  year={2021}
}

@inproceedings{guo2020single,
  title={Single path one-shot neural architecture search with uniform sampling},
  author={Guo, Zichao and Zhang, Xiangyu and Mu, Haoyuan and Heng, Wen and Liu, Zechun and Wei, Yichen and Sun, Jian},
  booktitle={European conference on computer vision},
  pages={544--560},
  year={2020},
  organization={Springer}
}

@article{banbury2021micronets,
  title={Micronets: Neural network architectures for deploying tinyml applications on commodity microcontrollers},
  author={Banbury, Colby and Zhou, Chuteng and Fedorov, Igor and Matas, Ramon and Thakker, Urmish and Gope, Dibakar and Janapa Reddi, Vijay and Mattina, Matthew and Whatmough, Paul},
  journal={Proceedings of Machine Learning and Systems},
  volume={3},
  pages={517--532},
  year={2021}
}

@article{krausz2019survey,
  title={A survey of teleceptive sensing for wearable assistive robotic devices},
  author={Krausz, Nili E and Hargrove, Levi J},
  journal={Sensors},
  volume={19},
  number={23},
  pages={5238},
  year={2019},
  publisher={MDPI}
}

@article{starke2022semi,
  title={Semi-autonomous control of prosthetic hands based on multimodal sensing, human grasp demonstration and user intention},
  author={Starke, Julia and Weiner, Pascal and Crell, Markus and Asfour, Tamim},
  journal={Robotics and Autonomous Systems},
  volume={154},
  pages={104123},
  year={2022},
  publisher={Elsevier}
}

@article{castro2022continuous,
  title={Continuous Semi-autonomous Prosthesis Control Using a Depth Sensor on the Hand},
  author={Castro, Miguel Nobre and Dosen, Strahinja},
  journal={Frontiers in Neurorobotics},
  volume={16},
  year={2022},
  publisher={Frontiers Media SA}
}

@article{jiang2021synergies,
  title={Synergies between affordance and geometry: 6-dof grasp detection via implicit representations},
  author={Jiang, Zhenyu and Zhu, Yifeng and Svetlik, Maxwell and Fang, Kuan and Zhu, Yuke},
  journal={arXiv preprint arXiv:2104.01542},
  year={2021}
}

@inproceedings{lin2022ondevice,
    title     = {On-Device Training Under 256KB Memory},
    author    = {Lin, Ji and Zhu, Ligeng and Chen, Wei-Ming and Wang, Wei-Chen and Gan, Chuang and Han, Song},
    booktitle = {Annual Conference on Neural Information Processing Systems (NeurIPS)},
    year      = {2022}
    }

@article{saha2022machine,
  title={Machine learning for microcontroller-class hardware-a review},
  author={Saha, Swapnil Sayan and Sandha, Sandeep Singh and Srivastava, Mani},
  journal={IEEE Sensors Journal},
  year={2022},
  publisher={IEEE}
}

@article{tang2022wearable,
  title={Wearable supernumerary robotic limb system using a hybrid control approach based on motor imagery and object detection},
  author={Tang, Zhichuan and Zhang, Lingtao and Chen, Xin and Ying, Jichen and Wang, Xinyang and Wang, Hang},
  journal={IEEE Transactions on Neural Systems and Rehabilitation Engineering},
  volume={30},
  pages={1298--1309},
  year={2022},
  publisher={IEEE}
}

@article{salminger2022current,
  title={Current rates of prosthetic usage in upper-limb amputees--have innovations had an impact on device acceptance?},
  author={Salminger, Stefan and Stino, Heiko and Pichler, Lukas H and Gstoettner, Clemens and Sturma, Agnes and Mayer, Johannes A and Szivak, Michael and Aszmann, Oskar C},
  journal={Disability and Rehabilitation},
  volume={44},
  number={14},
  pages={3708--3713},
  year={2022},
  publisher={Taylor \& Francis}
}

@article{sun2020real,
  title={Real-time radar-based gesture detection and recognition built in an edge-computing platform},
  author={Sun, Yuliang and Fei, Tai and Li, Xibo and Warnecke, Alexander and Warsitz, Ernst and Pohl, Nils},
  journal={IEEE Sensors Journal},
  volume={20},
  number={18},
  pages={10706--10716},
  year={2020},
  publisher={IEEE}
}

@article{chu2019learning,
  title={Learning affordance segmentation for real-world robotic manipulation via synthetic images},
  author={Chu, Fu-Jen and Xu, Ruinian and Vela, Patricio A},
  journal={IEEE Robotics and Automation Letters},
  volume={4},
  number={2},
  pages={1140--1147},
  year={2019},
  publisher={IEEE}
}

@article{chu2019toward,
  title={Toward affordance detection and ranking on novel objects for real-world robotic manipulation},
  author={Chu, Fu-Jen and Xu, Ruinian and Seguin, Landan and Vela, Patricio A},
  journal={IEEE Robotics and Automation Letters},
  volume={4},
  number={4},
  pages={4070--4077},
  year={2019},
  publisher={IEEE}
}

@article{xu2021affordance,
  title={An affordance keypoint detection network for robot manipulation},
  author={Xu, Ruinian and Chu, Fu-Jen and Tang, Chao and Liu, Weiyu and Vela, Patricio A},
  journal={IEEE Robotics and Automation Letters},
  volume={6},
  number={2},
  pages={2870--2877},
  year={2021},
  publisher={IEEE}
}

@incollection{ragusa2023affordanceAP,
  title={Affordance Segmentation Using RGB-D Sensors for Application in Portable Embedded Systems},
  author={Ragusa, Edoardo and Ghezzi, Matteo Pastorino and Zunino, Rodolfo and Gastaldo, Paolo},
  booktitle={Applications in Electronics Pervading Industry, Environment and Society: APPLEPIES 2022},
  pages={109--116},
  year={2023},
  publisher={Springer}
}

@article{ragusa2023affordance,
  title={Affordance segmentation using tiny networks for sensing systems in wearable robotic devices},
  author={Ragusa, Edoardo and Dosen, Strahinja and Zunino, Rodolfo and Gastaldo, Paolo},
  journal={IEEE Sensors Journal},
  year={2023},
  publisher={IEEE}
}

@article{xie2021segformer,
  title={SegFormer: Simple and efficient design for semantic segmentation with transformers},
  author={Xie, Enze and Wang, Wenhai and Yu, Zhiding and Anandkumar, Anima and Alvarez, Jose M and Luo, Ping},
  journal={Advances in Neural Information Processing Systems},
  volume={34},
  pages={12077--12090},
  year={2021}
}

@inproceedings{lugani2023lightweight,
  title={Lightweight Neural Networks for Affordance Segmentation: Enhancement of the Decoder Module},
  author={Lugani, Simone and Ragusa, Edoardo and Zunino, Rodolfo and Gastaldo, Paolo},
  booktitle={International Conference on Applications in Electronics Pervading Industry, Environment and Society},
  pages={437--443},
  year={2023},
  organization={Springer}
}

@inproceedings{deng20213d,
  title={3d affordancenet: A benchmark for visual object affordance understanding},
  author={Deng, Shengheng and Xu, Xun and Wu, Chaozheng and Chen, Ke and Jia, Kui},
  booktitle={proceedings of the IEEE/CVF conference on computer vision and pattern recognition},
  pages={1778--1787},
  year={2021}
}

@article{gao2024dense,
  title={Are Dense Labels Always Necessary for 3D Object Detection from Point Cloud?},
  author={Gao, Chenqiang and Liu, Chuandong and Shu, Jun and Liu, Fangcen and Liu, Jiang and Yang, Luyu and Gao, Xinbo and Meng, Deyu},
  journal={arXiv preprint arXiv:2403.02818},
  year={2024}
}

@article{zhou2024dgpinet,
  title={DGPINet-KD: Deep Guided and Progressive Integration Network with Knowledge Distillation for RGB-D Indoor Scene Analysis},
  author={Zhou, Wujie and Jian, Bitao and Fang, Meixin and Dong, Xiena and Liu, Yuanyuan and Jiang, Qiuping},
  journal={IEEE Transactions on Circuits and Systems for Video Technology},
  year={2024},
  publisher={IEEE}
}

@article{chiariotti2024future,
  title={The Future of Bionic Limbs: The untapped synergy of signal processing, control, and wireless connectivity},
  author={Chiariotti, Federico and Mamidanna, Pranav and Suman, Suraj and Stefanovi{\'c}, {\v{C}}edomir and Farina, Dario and Popovski, Petar and Do{\v{s}}en, Strahinja},
  journal={IEEE Signal Processing Magazine},
  volume={41},
  number={4},
  pages={58--75},
  year={2024},
  publisher={IEEE}
}

@article{mastinu2024explorations,
  title={Explorations of autonomous prosthetic grasping via proximity vision and deep learning},
  author={Mastinu, E and Coletti, A and van den Berg, J and Cipriani, C},
  journal={IEEE Transactions on Medical Robotics and Bionics},
  year={2024},
  publisher={IEEE}
}

@article{akkad2023embedded,
  title={Embedded deep learning accelerators: A survey on recent advances},
  author={Akkad, Ghattas and Mansour, Ali and Inaty, Elie},
  journal={IEEE Transactions on Artificial Intelligence},
  year={2023},
  publisher={IEEE}
}

@article{lopes2022survey,
  title={A survey on RGB-D datasets},
  author={Lopes, Alexandre and Souza, Roberto and Pedrini, Helio},
  journal={Computer Vision and Image Understanding},
  volume={222},
  pages={103489},
  year={2022},
  publisher={Elsevier}
}

@article{rana2024affordance,
  title={Affordance-Centric Policy Learning: Sample Efficient and Generalisable Robot Policy Learning using Affordance-Centric Task Frames},
  author={Rana, Krishan and Abou-Chakra, Jad and Garg, Sourav and Lee, Robert and Reid, Ian and Suenderhauf, Niko},
  journal={arXiv preprint arXiv:2410.12124},
  year={2024}
}

@article{song2024efficient,
  title={Efficient Evaluation Methods for Neural Architecture Search: A Survey},
  author={Song, Xiaotian and Xie, Xiangning and Lv, Zeqiong and Yen, Gary G and Ding, Weiping and Lv, Jiancheng and Sun, Yanan},
  journal={IEEE Transactions on Artificial Intelligence},
  year={2024},
  publisher={IEEE}
}

@article{li2024zero,
  title={Zero-Shot Neural Architecture Search: Challenges, Solutions, and Opportunities},
  author={Li, Guihong and Hoang, Duc and Bhardwaj, Kartikeya and Lin, Ming and Wang, Zhangyang and Marculescu, Radu},
  journal={IEEE Transactions on Pattern Analysis and Machine Intelligence},
  year={2024},
  publisher={IEEE}
}

@article{ozccil2024affordance,
  title={Affordance Labeling and Exploration: A Manifold-Based Approach},
  author={{\"O}z{\c{c}}il, {\.I}smail and Koku, A Bu{\u{g}}ra},
  journal={arXiv preprint arXiv:2407.15479},
  year={2024}
}

@article{wang2024sgsin,
  title={SGSIN: Simultaneous Grasp and Suction Inference Network via Attention-Based Affordance Learning},
  author={Wang, Wenshuo and Zhu, Haiyue and Ang Jr, Marcelo H},
  journal={IEEE Transactions on Industrial Electronics},
  year={2024},
  publisher={IEEE}
}

@article{capogrosso2023machine,
  title={A Machine Learning-oriented Survey on Tiny Machine Learning},
  author={Capogrosso, Luigi and Cunico, Federico and Cheng, Dong Seon and Fummi, Franco and others},
  journal={arXiv preprint arXiv:2309.11932},
  year={2023}
}

@article{burrello2023enhancing,
  title={Enhancing Neural Architecture Search with Multiple Hardware Constraints for Deep Learning Model Deployment on Tiny IoT Devices},
  author={Burrello, Alessio and Risso, Matteo and Motetti, Beatrice Alessandra and Macii, Enrico and Benini, Luca and Pagliari, Daniele Jahier},
  journal={IEEE Transactions on Emerging Topics in Computing},
  year={2023},
  publisher={IEEE}
}

@article{salehin2024automl,
  title={AutoML: A systematic review on automated machine learning with neural architecture search},
  author={Salehin, Imrus and Islam, Md Shamiul and Saha, Pritom and Noman, SM and Tuni, Azra and Hasan, Md Mehedi and Baten, Md Abu},
  journal={Journal of Information and Intelligence},
  volume={2},
  number={1},
  pages={52--81},
  year={2024},
  publisher={Elsevier}
}

\end{document}